\documentclass[journal]{IEEEtran}
\usepackage{cite}

\ifCLASSINFOpdf
    \usepackage[pdftex]{graphicx}
    \graphicspath{{../pdf/}{../jpeg/}}
    \DeclareGraphicsExtensions{.pdf,.jpeg,.png}
\else
   \usepackage[dvips]{graphicx}
   \graphicspath{{../eps/}}
  \DeclareGraphicsExtensions{.eps}
\fi
\usepackage[cmex10]{amsmath}
\usepackage{mathtools}
\usepackage{algorithmicx}
\usepackage{algpseudocode}
\usepackage[ruled,vlined]{algorithm2e}
\DeclarePairedDelimiter\abs{\lvert}{\rvert}%
\usepackage{geometry} 
\usepackage{array}

\ifCLASSOPTIONcaptionsoff
  \usepackage[nomarkers]{endfloat}
 \let\MYoriglatexcaption\caption
 \renewcommand{\caption}[2][\relax]{\MYoriglatexcaption[#2]{#2}}
\fi
\begin{document}
%
\title{NSANet: Noise Seeking Attention Network }
%
%
%

\author{Maryam Jameela,~\IEEEmembership{Member,~IEEE}
        and~Gunho Sohn~\IEEEmembership{Member,~IEEE}
\thanks{}
\thanks{}
\thanks{}}

%
%

\markboth{IEEE TRANSACTIONS ON GEOSCIENCE AND REMOTE SENSING, VOL. NO, 2022}%
{Shell \MakeLowercase{\textit{et al.}}: Bare Demo of IEEEtran.cls for Journals}
%



\maketitle

\begin{abstract}
LiDAR (Light Detection and Ranging) technology has remained popular in capturing natural and built environments for numerous applications. The recent technological advancements in electro-optical engineering have aided in obtaining laser returns at a higher pulse repetition frequency (PRF), which considerably increased the density of the 3D point cloud. Conventional techniques with lower PRF had a single pulse-in-air (SPIA) zone, large enough to avoid a mismatch among pulse pairs at the receiver. New multiple pulses-in-air (MPIA) technology guarantees various windows of operational ranges for a single flight line and no blind zones. The disadvantage of the technology is the projection of atmospheric returns closer to the same pulse-in-air zone of adjacent terrain points likely to intersect with objects of interest. These noise properties compromise the perceived quality of the scene and encourage the development of new noise-filtering neural networks, as existing filters are significantly ineffective. We propose a novel dual-attention noise-filtering neural network called Noise Seeking Attention Network (NSANet) that uses physical priors and local spatial attention to filter noise. Our research is motivated by two psychology theories of feature integration and attention engagement to prove the role of attention in computer vision at the encoding and decoding phase. The presented results of NSANet show the inclination towards attention engagement theory and a performance boost compared to the state-of-the-art noise-filtering deep convolutional neural networks.
\end{abstract}

\begin{IEEEkeywords}
systematic noise, computer vision, attention, noise filtering neural network, physical priors, point cloud 
\end{IEEEkeywords}

%
\IEEEpeerreviewmaketitle

\section{Introduction}
%
%
%
%
\IEEEPARstart{L}IDAR (Light Detection and Ranging) is a crucial sensor for mapping natural and built environments for defence and civil applications. Lately, technological advancements in electro-optical engineering have aided in obtaining laser returns at high pulse repetition frequency (PRF), which increases the density of the 3D point cloud considerably. Even though these lidars have been used to capture the reality of the scene, sometimes, the perceived quality of the scene is compromised by unwanted reality (e.g., atmospheric data). This unwanted atmospheric data is called noise \cite{OverView}.

\begin{figure}
\centering
\includegraphics[width=8.0cm]{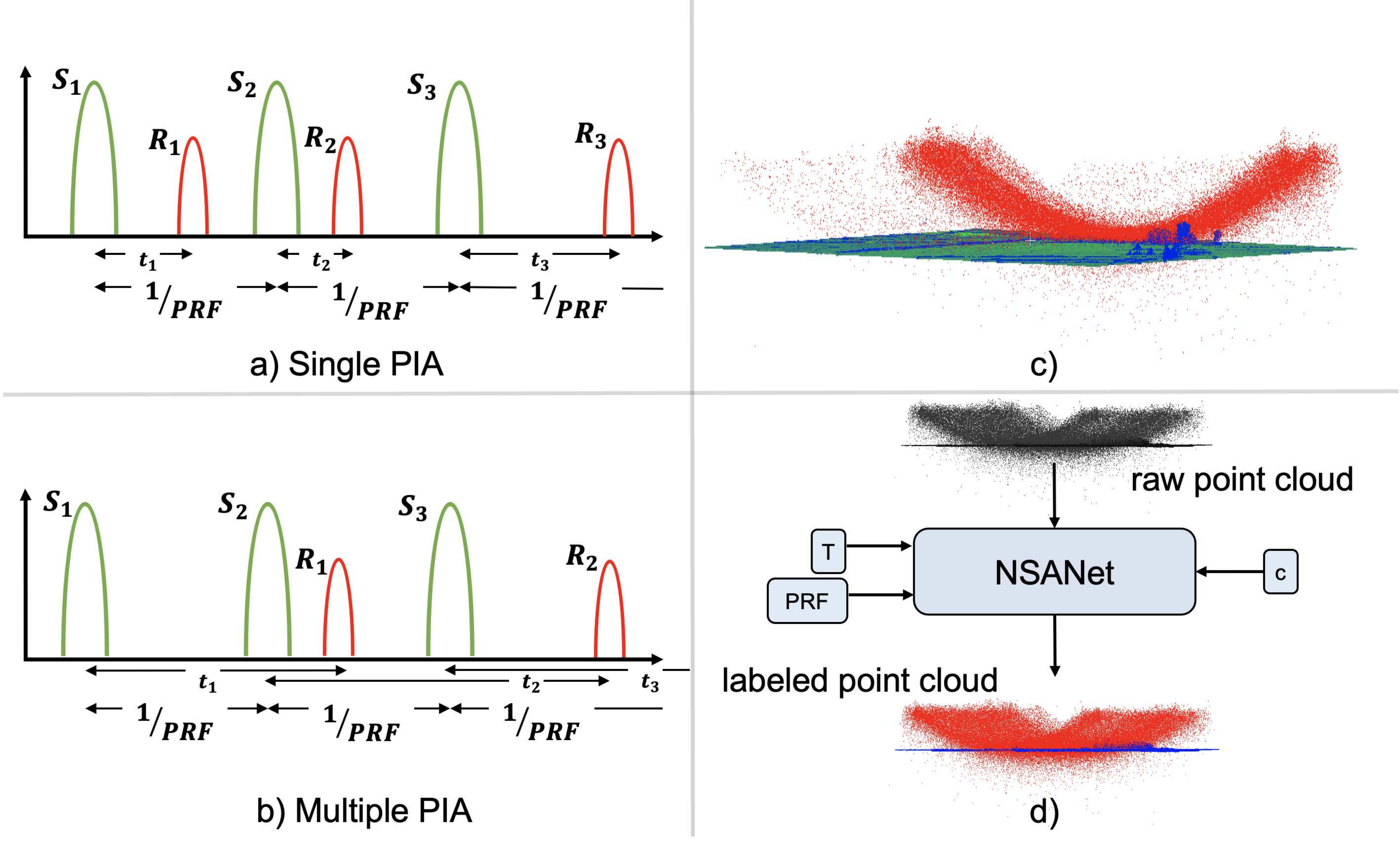}
  \caption {a) SPIA is a traditional technology which matched the signal sent by a transmitter (S) and received by a receiver(R) efficiently but was expensive, b) MPIA technology improved the cost and speed up the data collection though it is more prone to atmospheric points which can be seen overlapping with objects of interest in c), d) The proposed NSANet uses physical priors and spatial attention to filter noise from the point cloud.}
  \label{fig:fig1}
\end{figure}

Traditional lidar technology completes the survey in a single pulse-in-air (SPIA) zone with lower PRF. It easily matches laser pulse pairs at the receiver (R) as shown in Figure \ref{fig:fig1}-a. These systems are slow, expensive, and have single operational ranges per flight line. Multiple pulses in air (MPIA) technology was introduced in 2006 to overcome these limitations. New fast and cost-effective gate-less sensors ensure no blind zones by generating high-density point clouds, and provide multiple operational ranges for a single flight, as seen in Figure \ref{fig:fig1}-b. However, the shortcoming of this technology is its vulnerability to atmospheric points. The noise points are mostly projected closer to or overlap with objects of interest. Increasing PRF worsens the situation by introducing narrower PIA zones which often causes an overall increase in the number of noise points \cite{PointCleanNetRakotosaona2020}. Figure \ref{fig:fig1}-c shows the compromised perception quality of the scene due to noise.  \par
While the atmospheric returns always subsisted at lower PRF, we could eliminate them with simple height filtering, or nearest-neighboring algorithms \cite{ImageDeReviewGoyal2020}. However, more sophisticated filtering algorithms are required once noise begins to overlap with the key objects and compromise the scene’s semantics. \par
Deep convolution neural networks have recently outperformed traditional state-of-the-art computer vision techniques \cite{pointnet, pointnetplusplus, 3DUNET, att3DUnet}. They can mimic human vision for perceiving 3D scenes through complex and deep feature extraction to learning patterns for detection, classification, and semantic segmentation. This motivated us to develop a large-scale noise classification deep neural network for efficient and robust noise filtering from the point cloud. Our data analysis demonstrated characteristics that facilitated our system design. \par
Our data shows variable noise density exhibiting characteristics like complex objects (i.e., trees and bushes). Furthermore, the global systematic noise pattern is localized on the transition point between two PIA zones and overlaps with objects of interest. Other challenges are (i) imbalanced class distribution, i.e., noise points making up (1-3)\% of the complete scene (ii) lack of annotated diverse dataset. \par
Our proposed method, Noise seeking attention network (NSANet), is a novel dual attention noise filtering neural network. Our key contributions are:

\begin{itemize}
    \item A dual attention-based noise filtering neural network for the large-scale 3D point cloud.
    \item  Dual attention module, which uses physical priors and local spatial attention to guide the network to activate a global noise pattern and eliminate ambiguity between overlapping objects \cite{att3DUnet} as shown in Figure \ref{fig:fig1}-d.
    \item Ablation study focused on the integration of attention into a network based on two psychology studies proposed by Treisman in 1993 called Feature Integration Theory (FIT)  \cite{TREISMAN198097} and Duncan and Humphrey in 1989 called Attentional Engagement Theory (AET) \cite{duncan1992beyond}.
\end{itemize}
The following sections discuss related work, motivation, methodology, experimental configuration, and results.

\section{Literature Review}
Noise filtering techniques can be broadly divided into four categories: a) statistical and geometrical, b) clustering-based noise filtering, c) machine learning, and d) deep learning.
\subsection{Statistical and Geometrical Techniques}
Traditionally, there are multiple statistical techniques used for filtering noise which include estimating univariate location and scale, low-pass filters, mean filters, median filters, principal component analysis, multivariate location, and covariance estimation, and spatial filters \cite{robustNoiseFiltering,snowfiltering,denoisingSnow,Rousseeuw2011RobustSF,bilateralfiltering, PNoiseClass, MRPCA}. These filters are used for cleaning 3D point clouds and images. Such naive approaches were very helpful for the conventional single pulse in air technology for data collection. Due to noise density and complexity, these techniques are ineffective and expensive for large-scale 3D point clouds. \par
Geometrical techniques have always been used for outlier detection in 3D point cloud, such as density \cite{densitybasedNoise}, distance \cite{KDtree}, graph \cite{GDTPC,dinesh2018fast, dinesh20183d, BLFiltering}, and probability-based methods. Furthermore, they were mostly employed for denoising 3D meshes, models, and sparse point clouds. Meanwhile, these techniques were transformed into voxelization pipelines, and density or distance-based filters were applied to remove noise \cite{DEnoising3DGrid}. Bipartite graphs for approximating errors (noise) and manifold-to-manifold distance-based methods for outlier detection have shown good performance \cite{2021}.  \par
However, geometrical methods are also ineffective and expensive for large-scale 3D point clouds. There is substantial performance degradation for clustered and complex noise. The research work in \cite{reflectivenoisefiltering} deals with a large-scale point cloud that contains single-echo reflection values. It has dealt with multi-sensor-based point cloud to remove reflection-based noise. Nonetheless, it doesn't work for MPIA-based lidar due to multiple returns.
\subsection{Cluster-based Noise Filtering}
These are techniques based on clustering for outlier detection and filtering noise, such as the nearest neighbor \cite{Irfan20213DPC}, radius outlier removal \cite{OutlierD}, mean-shift clustering, K-Means \cite{robustNoiseFiltering}, and DBScan \cite{DBScan}. However, they require extensive memory for constructing clusters tight enough to preserve the object points and remove outliers and corrupt points. These practices work well for a dataset with low overlaps with objects of interest. Although, over time, they have been modified to run on GPUs for a performance boost, covering each point and performing recursive clustering is resource intensive. It's important to mention that most of these techniques are used by 3D point cloud models instead of large-scale scenes.
\subsection{Machine Learning Techniques}
The advent of machine learning has revolutionized data analysis procedures. It also came in handy for cleaning 3D point clouds. Different communities have employed linear regression, support vector machine \cite{SVM_denosing}, random forest \cite{IRF}, and many others for filtering or eliminating noise. Existing tools have been using these methodologies, successfully preserving the object's points while removing obvious noise. The trade-off between preserving the scene quality and removing noise depends on the application domain's analysis. However, They require feature crafting and selection to feed into the system for learning the pattern. Manual feature crafting and selection require extensive study of the dataset and domain. Hence, noise filtering researchers started exploring convolution neural networks after introducing deep learning for scene understanding.
\subsection{Deep Learning}
Deep learning in computer vision has significantly improved segmentation, classification, reconstruction, and detection. Convolution neural networks extract features from input data and detect decision-making patterns. They have shown performance boost in denoising, surface reconstruction, densification, noise classification, and filtering. PointCleanNet \cite{PointCleanNetRakotosaona2020}, PointFilter \cite{pfiltering}, Neural Project Denoising (NPD) \cite{NPD}, MaskNet \cite{masknet}, and CNN-based normal estimation \cite{PCUNE} for smoothing have improved filtering noise using various techniques through deep neural networks. Recently, transformer-based encoder-decoder has been used to filter noise within point clouds containing single-echo reflection values \cite{transformnerrefl}. It converts the point cloud into range images and removes noise. The work is efficient for single-echo reflective sensors but couldn't generate range images for airborne lidar using MPIA technology. \par
Most previous research is promising and shows potential for noise-filtering 3D point clouds. The limitation of the previous research work is the lack of a dedicated network for a large-scale 3D point cloud. These methods have not shown comparable performance on complex and clamped noises. Our novel deep neural noise filtering network exhibits the efficiency of handling large-scale point clouds and delivers a performance boost. It utilizes the physical priors of the dataset and embeds it as attention to the network for filtering noise points, further validating our hypothesis regarding embedded attention and physical priors. The next section will discuss attention and its use in computer vision.
\subsubsection{Attention}
\begin{figure}
\centering
\includegraphics[width=8.0cm]{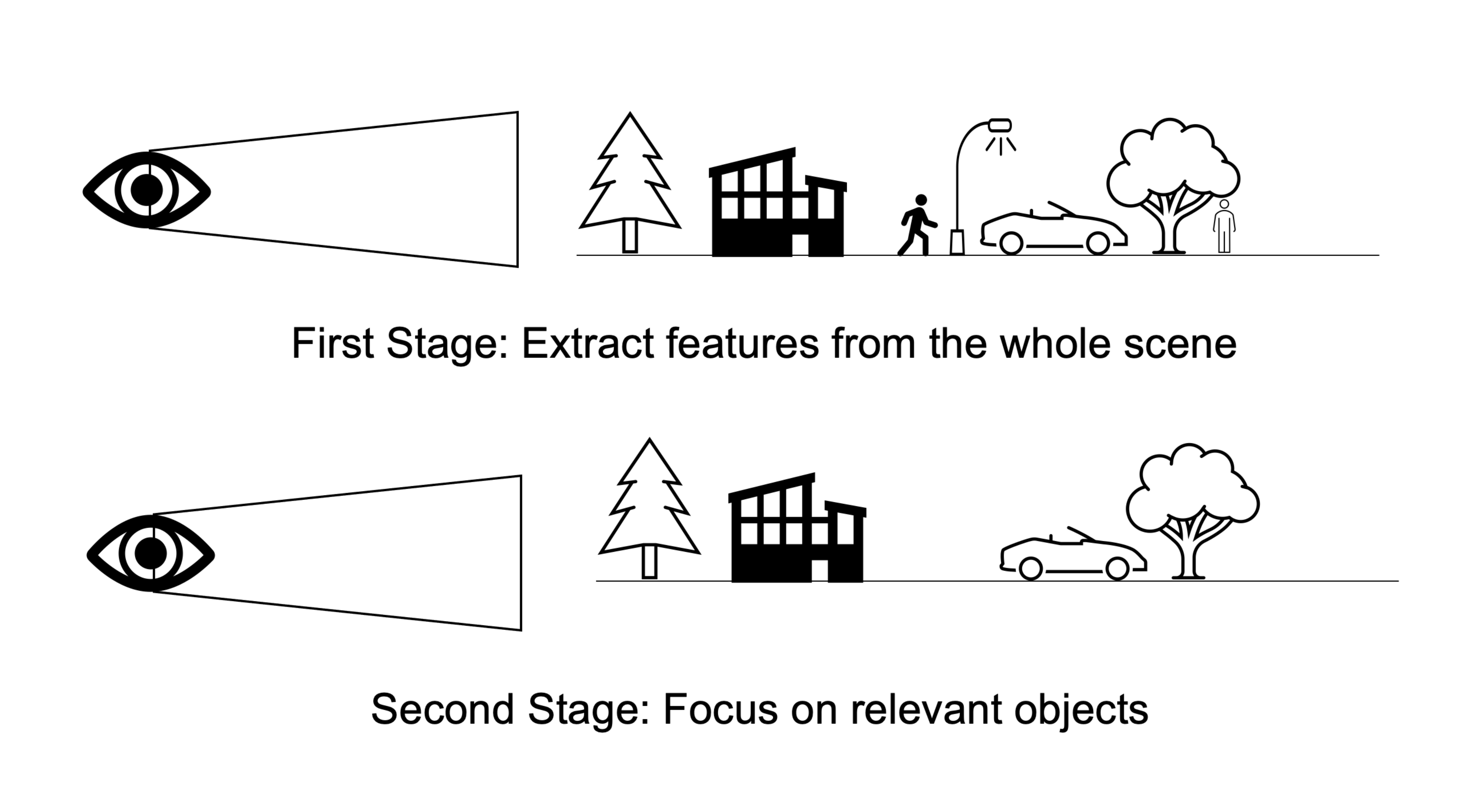}
  \caption{Attention phenomena and stages: First stage where low-level features are extracted, and a second stage where the focus is on relevant regions based on an earlier stage.}
  \label{fig:fig2}
\end{figure}

Attention is a phenomenon that encourages withdrawal from some things to focus on others. It is two-staged process shown in Figure \ref{fig:fig2}. The first stage focuses on extracting cues from the complete scene, while the second emphasizes relevant details. Attention is seen as the process of diverting focus on more important features for classification, object detection, segmentation, and object tracking. It is the process that has been adapted from human vision into computers for making better judgments and extracting significantly crucial features. Attention is all you need \cite{attenAllYouNeed} was the first publication that integrated the attention mechanism in NLP and motivated its use in other mainstream deep learning tasks across various fields, such as computer vision.
Over the years, multiple attention types such as single-head, multi-head, additive, multiplicative, spatial, channel, and fusion attention have been developed and studied \cite{att}. Where and what to attend have been a major research questions. Spatial self-attention showed performance improvements for segmentation and object detection problems and showed that additive self-attention could work for ambiguous boundaries between objects of interest \cite{att3DUnet}. The research explored the psychological theories of attention based on the following three-stage paradigms:
\begin{itemize}
\item A context-independent extraction of low-level features.
\item Modeling relationship and analysis of low-level features are based on the computer vision task.
\item Focus on relevant regions based on earlier stages of extracting high-level features for the final task \cite{Yeshurun1997AttentionalMI}.
\end{itemize}
\subsubsection{Attentional Engagement Theory (AET)} Duncan and Humphrey proposed AET in 1989 \cite{duncan1992beyond}. The theory divides the
attention process into two stages i) the pre-attentive stage and ii) selective attention. It presents the idea of extracting features from the data and then paying attention to features intensely and clearly for deciding. Regarding computer vision and deep learning, our articulation of a theory is the attention-free encoding of features and attention-based decoding. 
\subsubsection{Feature Integration Theory (FIT)} Treisman’s 1993 contradicted AET and proposed feature integration theory \cite{TREISMAN198097}. FIT proposed to pay attention to feature extraction and decide based on the selectively attentive feature. Research interprets this as an attention-based encoder and decoder to implement the FIT theory.

\section{Motivation}
Noise is a by-product of the latest technological advancements and surveying environment. Our proposed network utilized the sensor's physical properties and mapping principles to filter noise. Two essential studies were conducted to design the network, i) data analysis and ii) physical priors. \BlankLine
\subsubsection{Data Analysis}
The first study identified the essential characteristics of a dataset. The following sections will discuss the findings and key features of the study:
\begin{itemize}
\item {Density:} Figuring out variable noise point density throughout the scene is crucial for the network design due to the ambiguous boundary between noise and foreground. Some of the noise is clustered, exhibiting the characteristics of complex objects such as trees or bushes, while other noise points are sparse.
\item{Localization:} Another common discriminating feature observed in the dataset is the localization of noise points. As discussed in previous sections, most of the noise points are projected along with the transition point between two PIA zones. These atmospheric points develop a systematic global pattern that could be dense or sparse, depending upon the environmental conditions of the surveying area. In some cases, this global recurring noise pattern is close to the objects of interest, while it is found below the terrain for others. 
\item{Distribution:} PIA zones are narrower due to higher PRF, giving rise to the overlapping atmospheric points projection to the foreground. These overlaps are commonly found with trees, bushes, buildings, powerlines, and sometimes with the terrain. Atmospheric points are few (1-3\%) compared to the dataset's foreground.
\end{itemize} \BlankLine
\subsubsection{Physical Priors}
\begin{figure}
\centering
\includegraphics[width=8.0cm]{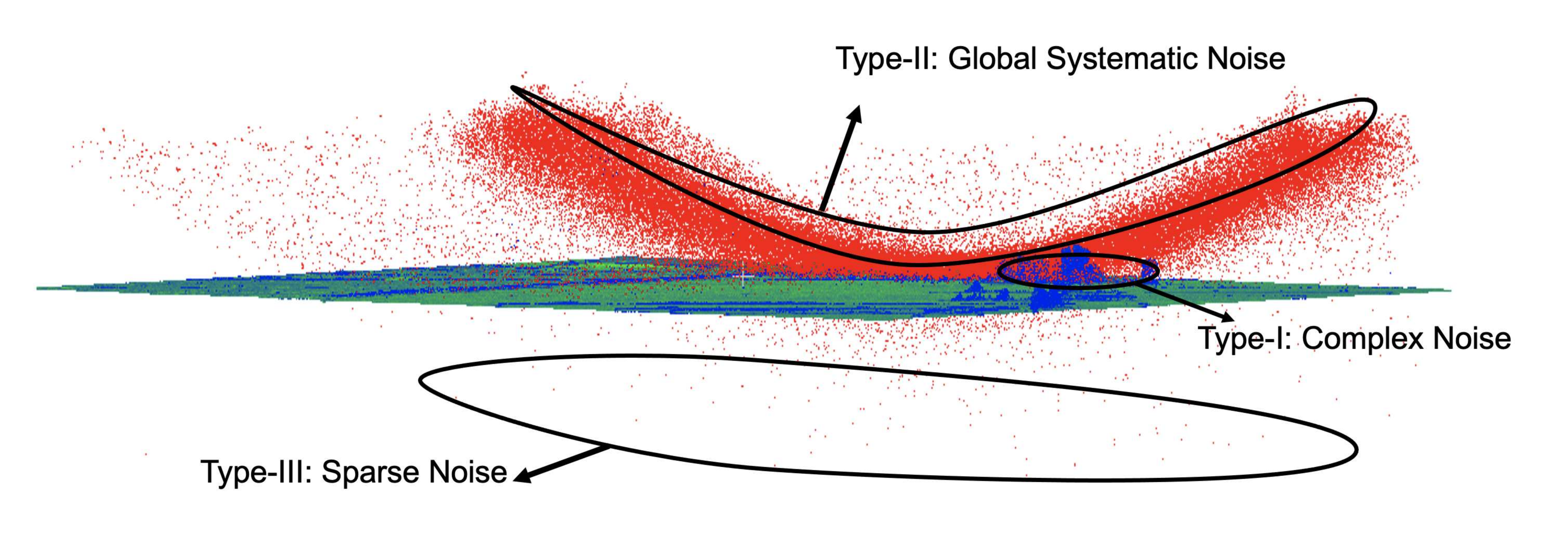}
  \caption{The three noise types: (1) Complex, (2) Global Systematic, and (3) Sparse Noise.}
  \label{fig:fig3}
\end{figure}

Electro-optical engineering and multiple pulses in air technology give a decent overview of the mapping of 3D scenes. We performed a comprehensive analysis to understand the distribution of noise data points, projection of noise points near the PIA transition zone as a systematic parabola, complex clustered noise points overlapped with objects of interest, and their characteristics of compromising the perceived quality of the scene \cite{SIM}. Our network takes advantage of these physical priors, which can help calculate the proximity of each point from the PIA transition zone. This proximity data helps generate a heatmap of the probability of a point being a noise. Heatmap is later used in our network as an external cue or physical priors to better incorporate global context for filtering noise.
Based on these studies, noise is divided into three types shown in Figure \ref{fig:fig3}, which also motivated to propose a design based on embedding physical priors to filter noise.
\begin{itemize}
    \item Complex Noise: Noise shows similar characteristics to trees and bushes. It is clustered and overlaps with objects of interest.
    \item Global Systematic Noise: Parabolic systematic-pattern dense or sparse localized at the transition point between PIA-zones refer to Figure \ref{fig:fig3}.
    \item Sparse Noise: Outlier and sparse noise points located far from terrain and objects of interest.
\end{itemize}
\section{Methodology}
NSANet is a multi-resolution dual attention-based noise filtering deep neural network. We propose to focus on all three types of noise in our network. Also, we reduce the ambiguity of boundaries between noise and non-noise objects through spatial and physical priors-based attention \cite{att3DUnet}. We will discuss the importance of the placement of the attention module in encoder and decoder blocks based on AET \cite{duncan1992beyond} and FIT \cite{TREISMAN198097} theories.
\begin{figure*}
\centering
\includegraphics[width=15.0cm]{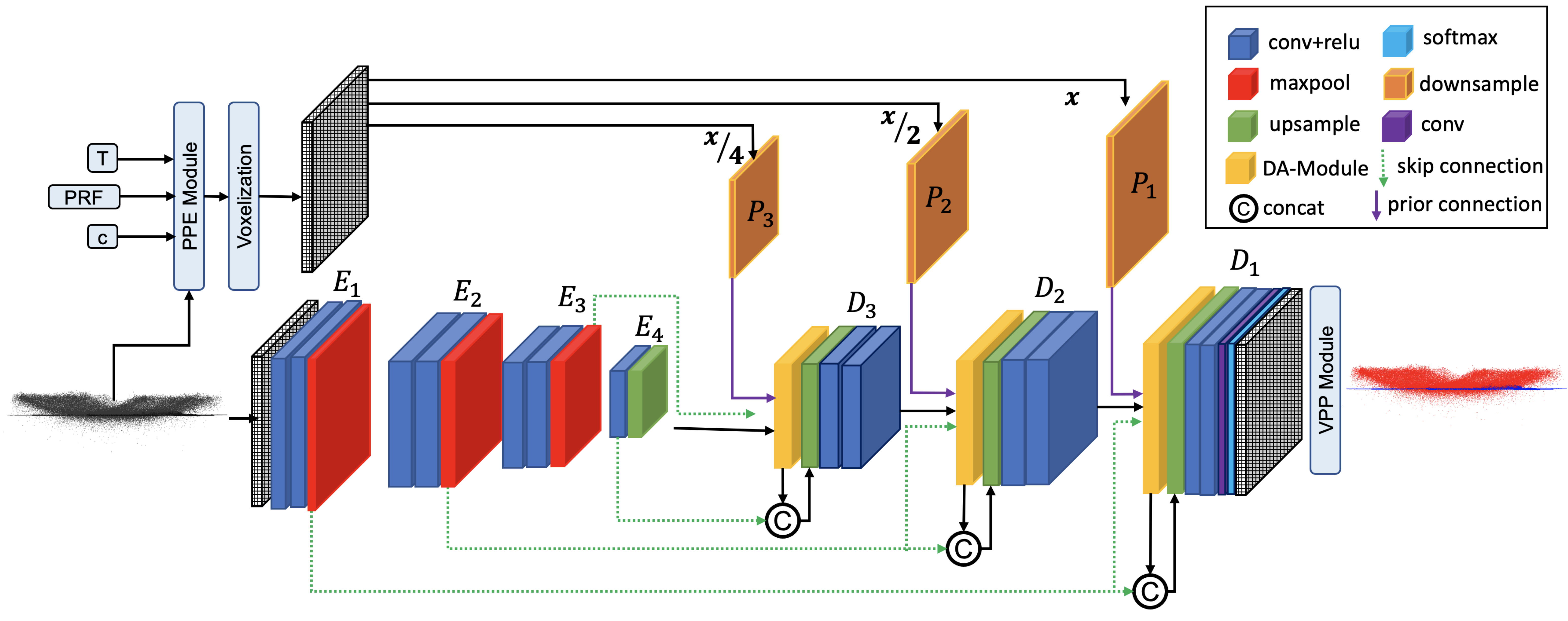}
  \caption{NSANet based on attentional engagement theory. It uses physical priors estimation (PPE) module to generate a PIA proximity probability map converted into a voxel grid and downsample by a factor of 2. Multiresolution encoder-decoder uses a dual-attention (DA) module to eliminate noise based on local and global contexts. The voxel post-processing (VPP) module eliminates borderline misclassification of noise points. }
  \label{fig:nsanet}
\end{figure*}
\subsection{3D Noise Filtering Neural Network}
The NSANet architecture is shown in Figure \ref{fig:nsanet}. Noise filtering deep neural network is a U-shaped multi-resolution encoder-decoder. It learns deep features from a three-dimensional voxel grid to label, and filter noise and non-noise \cite{3DUNET}. Our selection of voxel-based architecture was influenced by a comparative study, which shows that 3D voxel-based networks have elevated performance and efficiency for all noise types as compared to point-based networks (PointNet \cite{pointnet}, and PointNet++ \cite{pointnetplusplus}). Our network encodes four feature maps of different resolutions \{{$E_1$}, {$E_2$}, {$E_3$}, and  {$E_4$}\}. A deeper layer extracts more sophisticated features, while the first few layers extract prominent low-level features. As shown in Figure \ref{fig:nsanet} {$E_n$} decodes feature map {$D_n$} with a series of operations, including local spatial additive attention applied on ${D_{n-1}}$ and $E_n$. Then, multiplicative attention is applied on the physical prior probability map $P_n$ as global attention and two 3x3x3 convolution operations followed by batch normalization and relu activation. It results in three output feature maps \{ $D_1$ , $D_2$ , $D_3$\} with dimensions ${\frac{H}{2l}\times\frac{W}{2l}\times\frac{D}{2l}\times {32l}}$ and $l$ is number of layer.  These decoded feature maps learn to focus on all three types of noise at multiple-receptive fields, integrate the global receptive through global attention to the label, and consequently remove noise points.
\subsubsection{Physical Priors Estimation Module}
The attention module requires a physical prior to estimating global attention. This module takes the speed of light $c$ and $PRF$ to estimate the maximum LIDAR range {$R_{max}$}. The system also takes maximum and minimum GPS time $P_{mingt}, P_{maxgt}$ recorded from the 3D point cloud to find the location $T_{match}$ of the sensor during the specific time. The trajectory of flight $T_m$ and 3D point clouds $P_n$ are used to estimate the range of each shot $R_n$. Then, {$R_{max}$} and {$R_n$} are used to estimate their respective PIA zones $PIA_n$. Additionally, PIA zones, $R_n$, and $R_{max}$, are used to calculate the observed range $R_{obs}$ beyond PIA zones which is translated into a probability map based on the distance from the transition point $Prob_{n}$.
\begin{algorithm}
\SetAlgoLined
\KwResult{$Prob$: Physical Priors Estimation Module: }
 \BlankLine
 $R_{MAX}=\frac{c}{2 \times PRF}$\;
 \BlankLine
 \While {$l \leq N_{T}$}{
    $T_{match} = T_l \leq P_{mingt}$ \ and \ $  T_l \geq P_{{maxgt}}$}
    
 \While{$n \leq N$}{
    $T_{m}= \abs(P_{gt}-T_{match})< 1e-2 $ \; 
    \BlankLine
    $R_n= \sqrt{({x_p}-{x_m})^2+({y_p}-{y_m})^2+({z_p}-{z_m})^2}$\;
    \BlankLine
    $PIA_{n}=ceil(\frac{R_{n}}{R_{max}})$\;
    \BlankLine
    $R_{obs}^{[n]}= R_{n} mod R_{max}$\;
    \BlankLine
    $Prob_{n}= R_{obs}/R_{max}$\;
 }
 \BlankLine
 \caption{Physical Priors Estimation Module.\newline
 \textbf{Input:} $PRF$, $c$, $P_{n}$, $T_{l}$, $N$, $N_{T}$\newline $n=1,2,3,.....,N, l=1,2,3,......,N_{T}$\newline$P_{mingt}, P_{maxgt}$}
\end{algorithm}

 \begin{tabbing} 
where \hspace{0.3cm} \= $N$ = Total no. of raw input points\\
\>$N_{T}$= Total no. of shots in flight trajectory\\

\end{tabbing}

\subsubsection{Dual Attention Module}
Sparse noise can be easily labeled by any shallow machine learning architecture or DAE \cite{DAE}. Complex noise tends to mix and exhibit similar characteristics as objects of interest due to the distribution of points and shape variance. The ambiguous boundaries and multi-receptive field can result in an overpopulation of false positives. The local additive attention gate handles these spatial ambiguities, reduces misclassification, and suppresses irrelevant features for labeling complex noise. \par
Global systematic noise patterns have been one of the most challenging problems, but it was noticed after careful study that it only occurs close to the PIA transition zone. We studied the physical priors, estimated the range and distance from the PIA transition zone, calculated the probability, and integrated it as global multiplicative attention. This attention focused on global systematic noise patterns.
These attention gates take input from the previous layer of decoder $D_1$ for $D_2$. Encoder feature map of the same level $E_2$ and prior probability map of the same level $P_2$ and generate attention co-efficient $\alpha$ to take element-wise summation with feature map $E_2$ as shown in the following Figure \ref{fig:figdam}.

\begin{figure}
\centering
\includegraphics[width=8.0cm]{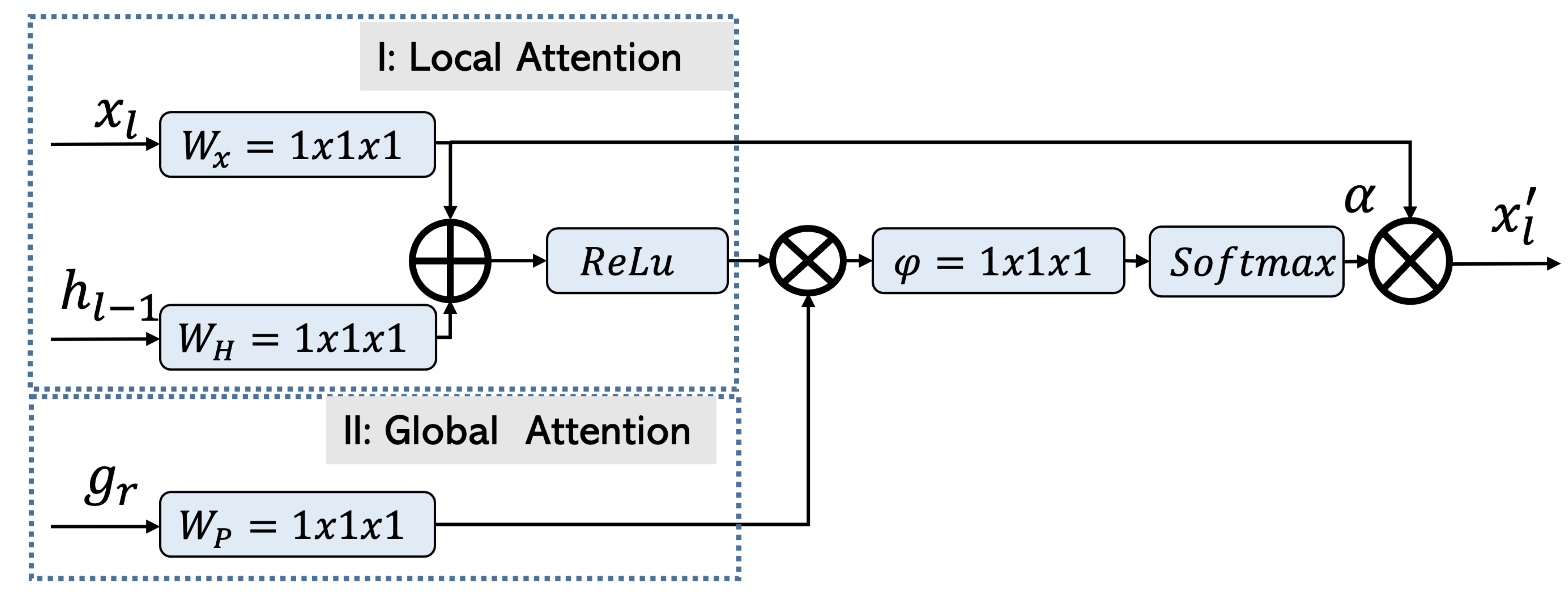}
  \caption{Dual attention module calculates attention coefficient based on the local and global feature map. Local feature map from the encoder and previous layer of the decoder and global feature maps comes from physical priors, which are calculated based on the global positioning of each point in the point cloud from the PPE module.}
  \label{fig:figdam}
\end{figure}
\subsubsection{Voxelization}
The input representation of the 3D point cloud for the network is a voxel grid. We preprocess the raw point cloud onto the voxel grid and calculate a mean representing all the points residing in the 3D voxel. The network maintains a projection matrix from the voxel grid to the raw point cloud for easy label projection. This voxel grid provides a trade-off between efficiency and effectiveness based on the selection of voxel size. 

\begin{figure*}
\centering
\includegraphics[width=15.0cm]{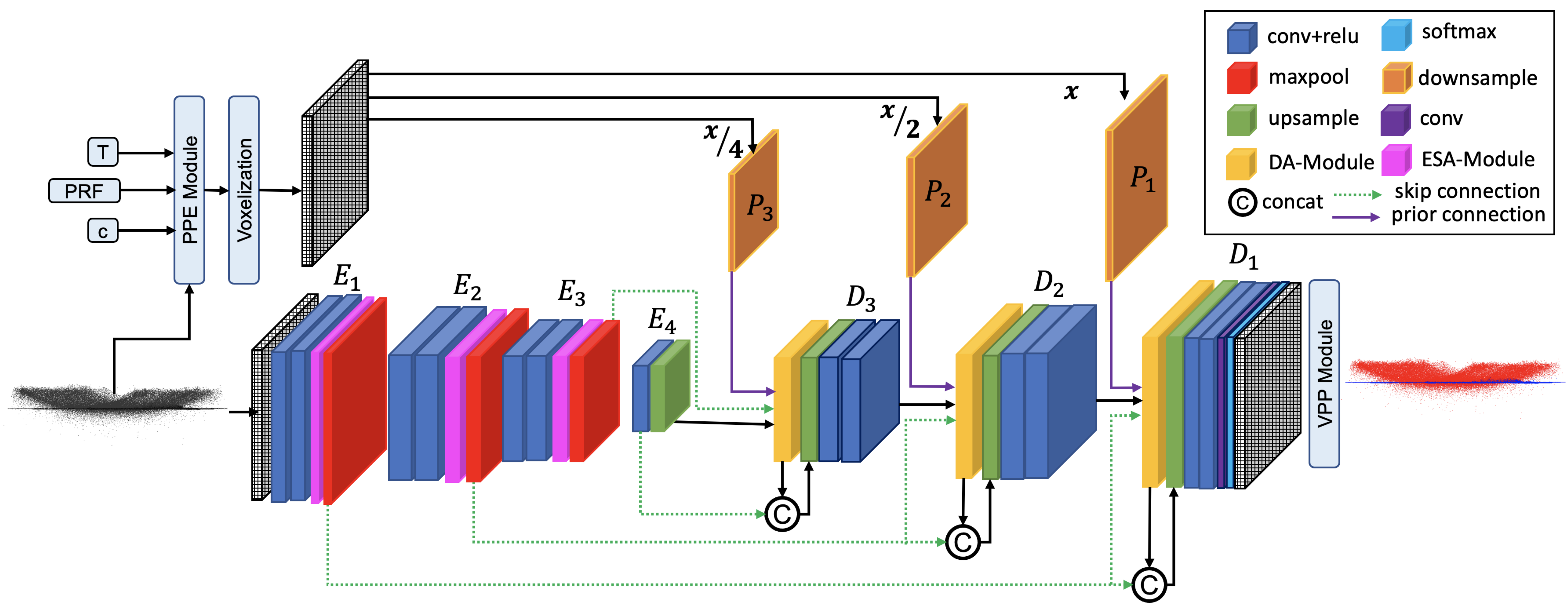}
  \caption{NSANet adapted feature integration theory by employing attention module in encoder as well. The first variation contains only the encoder module's local attention (FIT V-1).}
  \label{fig:fig6}
\end{figure*}
\begin{figure*}
\centering
\includegraphics[width=15.0cm]{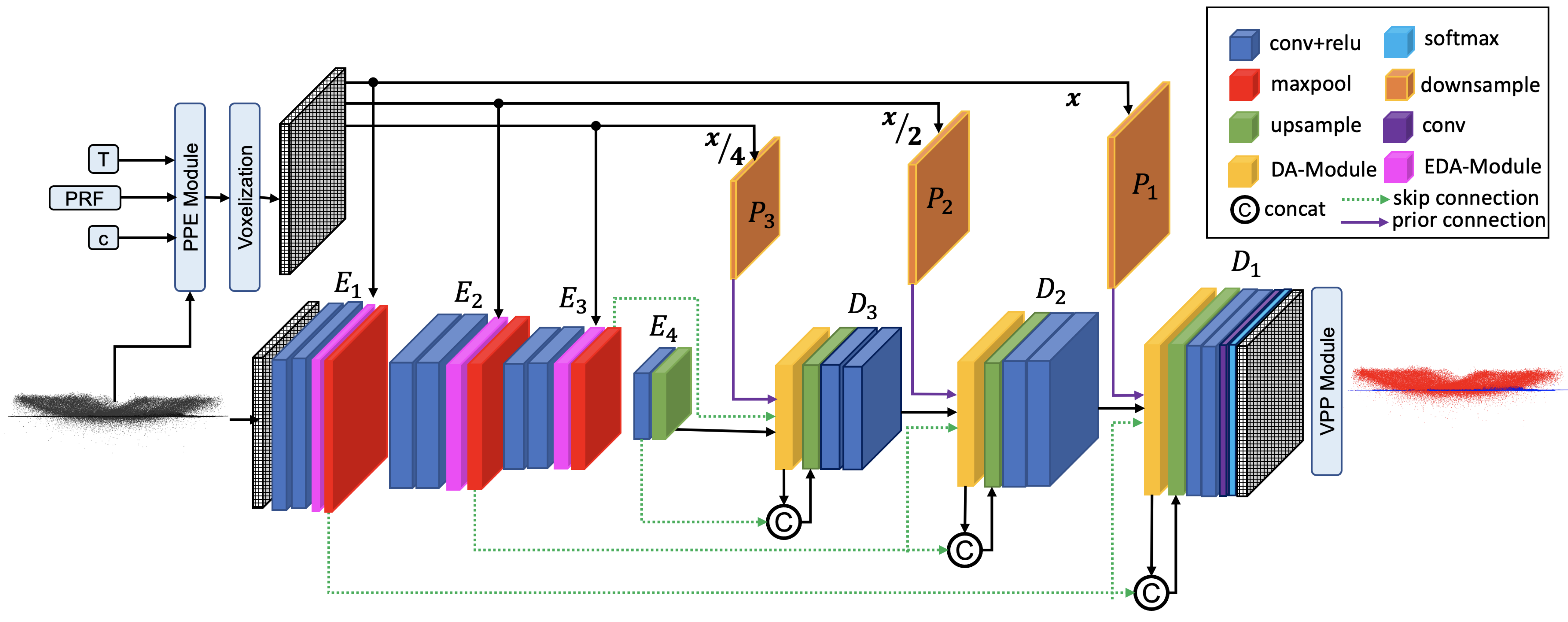}
  \caption{NSANet adapted feature integration theory by employing attention module in encoder as well. The second variation contains the encoder module's dual attention (FIT V-2).}
  \label{fig:fig7}
\end{figure*}

\section{Experiments and Results}\label{experiment} We conducted ablation studies based on feature engineering, and psychological theories of attention are used to plug in and out the attention modules. Our research also performed a comparative analysis and validated our choice of a voxel-based network for noise-filtering neural networks. The experiments assess test sets on the noise class. 
\subsection{Dataset}
Data is collected using a Teledyne Galaxy T1000 laser scanner over $13km^2$. Data is later divided into train and test sets for experiments. The first $4km^2$ of the dataset is used for testing, and the remaining data is used for training the network. There are 14 non-overlapping scenes containing more than five million points with an average density of $300 pp/{m^2}$. Teledyne Optech labeled our ground truth using Terrasolid point cloud processing software \cite{terrasolid}. Labels were generated through extensive domain knowledge and technical expertise. Our training dataset had two classes noise and non-noise. 
\subsection{Experimental Configuration}
Each scene is divided into four sub-scenes by GPS time of flight line. Our voxel grid is generated over each sub-scene, and the input grid size = $64m\times 64m \times 64m$ and Batch size = 64 grids with a voxel size of $1m^3$. Feature channels consist of mean absolute elevation and number of occupancy points. We have conducted a feature engineering study to select these features, which will be discussed in the following section. NSANet outputs $64m \times 64m \times 64m \times 2$, a confidence score against 3D classes (noise and non-noise). Later, the final prediction assigns a true label based on the highest confidence score, and post-processing is used to eliminate border-line misclassification and refine the prediction better.
NSANet is trained on two GPU RTX 6000 for 90 epochs; training time is between 48-60 hours, and inference takes about 2-3 mins.
\subsection{Evaluation Matrices}
We selected precision, recall, and F1 scores to evaluate our network's performance as noise in the minority class. The precision determines the total number of predicted noise or non-noise out of the total instances of the class. Recall measures correctly predicted samples out of the total predicted sample of the specific class. The F1 score shows the balance between precision and recall. 
\subsection{Feature Engineering}
The input features for deep learning networks are crucial for learning.  We initially selected a palette of features, including occupancy points (Occ.) for baseline 3D UNet. The mean elevation for each voxel is calculated based on the z values of each point residing in the voxel. It boosts performance by helping the network learn the vertical geometry of the scene, as reflected in Table \ref{tab:fe}. The third experiment added Occ., mean absolute elevation (MZ), num of returns (NR), intensity (INS), and range (R) probability. It performed better than the UNet, with only one or two channels. Though, it increased the size of the input immensely. The proposed network NSANet uses physical priors and spatial attention. Hence, we trained it with just one channel, and it outperformed 3D UNet. Then, We added mean elevation (MZ) to feature channels along Occ. The final results in Table \ref{tab:fe} show that absolute elevation is an important feature, especially because the noise distribution in elevation is very distinctive and discriminating for the neural network.
\begin{table*}[h]
  \begin{center}
	\centering
		\begin{tabular}{p{2.0cm}|p{3.5cm}|p{2.0cm}|p{2.0cm}|p{2.0cm}}\hline
			{Methods}&{Features} & {Recall} &{Precision} &{F1 Score}\\\hline
			 3D UNet & Occ & 68.5\% & 91.4\% & 0.783\\\hline
			 3D UNet & Occ + MZ & 72.5\% & 85.4\% & 0.789\\\hline
			 3D UNet & Occ+NR+INS+MZ+R & 82.7\% & 83.5\% & 0.831\\\hline
			 NSANet  & Occ & 77.9\% &91.4\% &0.841\\\hline
			 \textbf{NSANet}  & \textbf{Occ+ MZ} & \textbf{87.3\%} &\textbf{92.9\%}&\textbf{0.900} \\\hline
		\end{tabular}
		\BlankLine
\caption{Ablation Study on importance of feature engineering over recall, precision, and f1 score; Occ: No. of Occupancy Points, MZ: Mean Elevation, INS: Intensity, R: Range and NR: Number of Returns}
	\label{tab:fe}
	\end{center}
\end{table*}

\subsection{Loss Functions}
We initially used cross-entropy (CE) loss, a common choice for segmentation problems, but our problem required a weighted cross entropy (WCE). Our dataset has a huge class-imbalanced distribution, and assigning weights was very important. Our class weights helped and improved the performance drastically, as shown in Table \ref{tab:losstb}. We did test focal loss (FL) to see if the performance improved, but WCE outperformed by a great margin.
\begin{table*}[h]
  \begin{center}
	\centering
		\begin{tabular}{p{1.2cm}|p{1.6cm}|p{1.0cm}|p{1.0cm}|p{1.2cm}}\hline
			{Methods}&{Loss Function} & {Recall} &{Precision} &{F1 Score}\\\hline
			 3D UNet & CE & 64.0\% & 99.7\% & 0.781\\\hline
			 3D UNet & FL & 66.0 \% & 98.7\% & 0.795\\\hline
			 3D UNet  & WCE & \textbf{77.9\%} &\textbf{91.4\%}&\textbf{0.841} \\\hline
		\end{tabular}
	\label{tab:trainloss}
	\BlankLine
\caption{Ablation Study on importance of loss function over F1-Score; WCE: Weighted Cross-Entropy and HLC: Hierarchical Layout Consistency Loss}
	\label{tab:losstb}
	\end{center}
\end{table*}
\subsection{Attention-based Ablation Study}
Our ablation study plugs a dual attention module in the encoder and decoder based on our interpretation of FIT and AET. We verified through experiments that the AET implementation shown in Figure \ref{fig:nsanet}, which focuses on using attention at the decoding stage, outperforms the variations of FIT-based NSANet shown in Figures \ref{fig:fig6} and \ref{fig:fig7}. It shows that the network takes advantage of attention theory AET and experimentally validates its usability. Our results in Table \ref{tab:abfe} show that attention engagement at the stage of interpreting features to filter noise is a better strategy for this particular problem.

\begin{table*}[h]
  \begin{center}
	\centering
		\begin{tabular}{p{2.5cm}|p{1.8cm}|p{1.8cm} |p{1.8cm}|
		p{1.8cm}}\hline
			{Methods} & Theory & Recall & Precision & F1 Score \\\hline
			 NSANet  \textbf{(ours)} & AET  	&87.30\%	& 92.90\%	& 0.900    \\\hline
			 NSANet  \textbf{(ours)} & FIT  V-1	&83.31\% &	93.60\%&	0.882  \\\hline
			 NSANet  \textbf{(ours)} & FIT V-2	&92.12\% &	79.69\%&	0.856  \\\hline
			 \textit{\textbf{NSANet+ VPP \textbf{(ours)}} } & AET	&\textit{\textbf{98.90\%}}	& \textit{96.10\%}	& \textit{\textbf{0.975}}    \\\hline
		\end{tabular}
\BlankLine
	\caption{Ablation Study based on AET and FIT attention theory and compared performance over recall, precision, and f1-score.}
	\label{tab:abfe}
	\end{center}
\end{table*}
\begin{table*}
  \begin{center}
	\centering
		\begin{tabular}{p{3.5cm}|p{1.9cm}|p{1.9cm}| p{1.9cm}| p{1.9cm}}\hline
			 Methods & input &Recall & Precision & F1 Score \\\hline
			 SVM (benchmark) & segment& 46.78\% &	54.05\%&	0.501  \\\hline
			 DAE \cite{DAE} & voxel	&53.57\% &	\textbf{99.90\%}&	0.697  \\\hline
			 PointNet\cite{pointnet} & point	&68.29\% &	98.19\%&	0.805  \\\hline
			 PointNet++\cite{pointnetplusplus} & point  	&57.65\% &	90.42\%&	0.740 \\\hline
			 3D UNet \cite{3DUNET} & voxel	&77.92\%	& 91.43\%	& 0.841    \\\hline
			 Att-UNet \cite{att3DUnet}& voxel 	&80.51\%	& 92.00\%	& 0.862    \\\hline
			 Multiview 2D-UNet & views 	&80.00\%&92.90\%& 0.859 \\\hline
			 \textit{\textbf{NSANet \textbf{(ours)}} }& voxel	&\textit{\textbf{87.30\%}}	& \textit{92.90\%}	& \textit{\textbf{0.900}}    \\\hline
		\end{tabular}
		\BlankLine
	\caption{Comparison of NFANet(ours) for noise filtering with SVM, Denoising Autoencoder,3D UNet, PointNet++ and PointNet over recall, precision and f1-score for noise class.}
	\label{tab:compartive}
	\end{center}
\end{table*}
\begin{figure*}
\centering
\includegraphics[width=16.0cm]{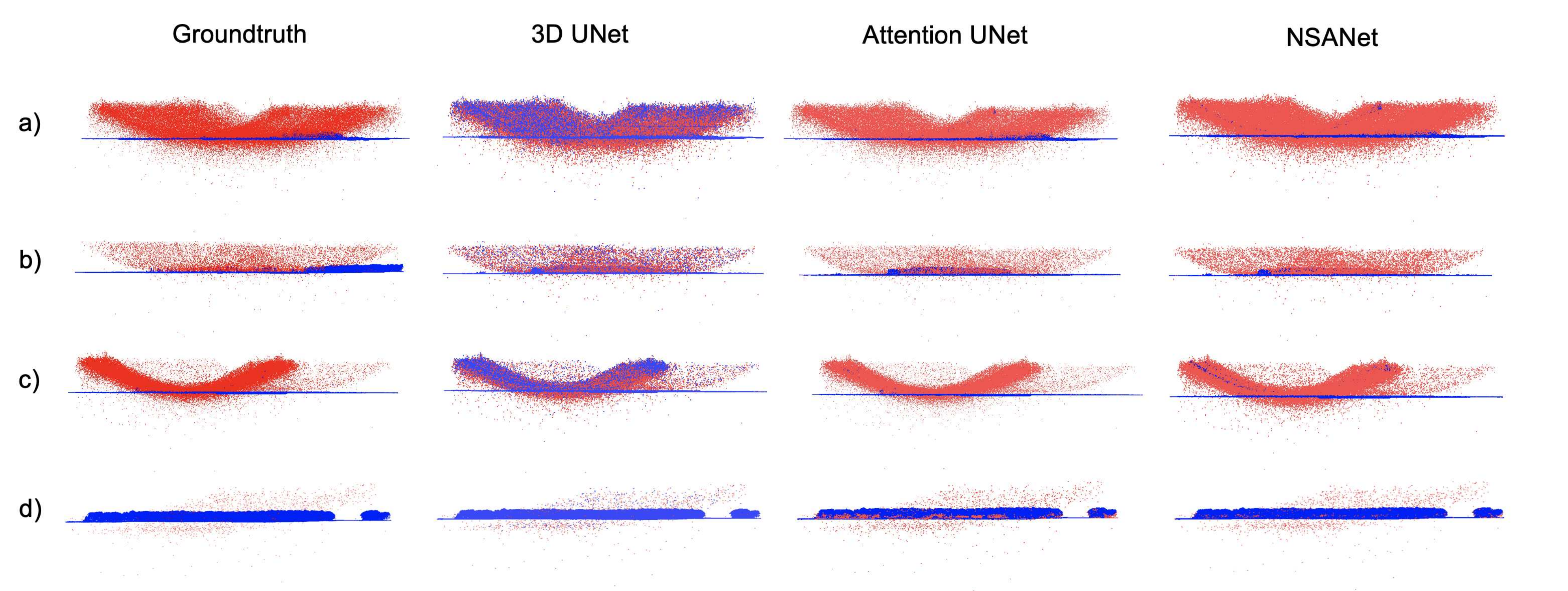}
  \caption{Visualization shows four different scenes comparing the SOTA performance of 3D UNet with cross-entropy loss, Attention UNet with weighted cross-entropy loss, and NSANet variation employed AET and weighted cross-entropy loss. Results reflect NSANet's superior performance. Red shows nose points, and blue reflects non-noise points.}
  \label{fig:fig8}
\end{figure*}
\subsection{Comparative Study}
We also developed a comparative study based on the literature review and chose SVM, DAE, PointNet, PointNet++, 3D UNet, Attention 3D UNet, and Multi-view-based 2D UNet.
\begin{itemize}
    \item  SVM was trained to find the trade-off between preserving objects of interest and eliminating noise points. It was able to achieve a 0.501 f1 score for the noise class.
    \item Denoising autoencoder (DAE) reconstructs the scene by removing noise points. SVM was noise classification while DAE was denoising reconstruction, showing comparatively better results of 0.697 f1 score on noise class.
    \item Pointnet \cite{pointnet}, a pioneering network to use points directly from the 3D point cloud. It performed better than DAE but failed over complex noise and lacked the global context, multi-resolution, and local contextual features.
    \item We also tested pointnet++ \cite{pointnetplusplus}, and it failed to deliver optimum results as well.
    \item Next, we selected 3D UNet with weighted cross entropy loss \cite{3DUNET} as our baseline, which used a multi-resolution encoder-decoder to include more context and showed exponentially good performance on noise class.
    \item We compared the performance with attention to 3D UNet and multiview-based 2D UNet. 2D UNet takes two different views as input and shows comparable performance to attention 3D UNet.Results in Figure show that Attention UNet\cite{att3DUnet} is aggressive and classifies non-noise as noise, while 3D UNet with cross-entropy loss performs poorly generally.
    
\end{itemize}      
Our conclusions from these experiments helped us to design our system, which utilizes global and local context, embedded it as attention to network, and showed impressive performance boost, as shown in Table  \ref{tab:compartive}. Our model maintained a performance balance across all the metrics, unlike the other networks, which have better precision but a lower recall and F1.

\subsection{Error Analysis}
Our results have demonstrated tremendous improvement in noise filtering compared to SOTA, as shown in Figure \ref{fig:fig8}. It is to be noted that our work takes inspiration from semantic segmentation and denoising research. Noise filtering from large-scale 3D point cloud for airborne lidar is a problem with very little work done to resolve it. Our network does struggle to filter in some areas, especially where the distribution of objects of objects-of-interest is very similar to noise or overlaps with noise. These cases are hard examples and create a performance gap. The voxel-based post-processing (VPP) module helps improve recall by 10\% based on the confidence of neighboring voxels as reflected in Table \ref{tab:abfe} .

\section{Conclusion}
This work presented a novel dual-attention NSANet that uses physical priors and local spatial attention to filter noise. Our experimental ablation study showed that the global and local context embedded as the attention could help deal with the ambiguity of overlapping objects and remove the systematic global pattern and sparse noise. It also proved that global context could help understand the scene better. These experiments have verified our conceptual inspiration and hypothesis of two-stage-based attentional engagement theory to integrate physical priors into the network. It showed our hypothesis not only confirms but also boosts noise filtering performance. NSANet demonstrated its capacity and efficiency to deal with the large-scale point clouds obtained from airborne LiDAR scanners. Our model performed consistently across all the metrics, unlike others. Our future work will focus on point-based neural networks and physical priors learning to remove noise.


%

\section*{Acknowledgment}
This research project has been supported by the Natural Sciences and Engineering Research Council of Canada
(NSERC)’s Collaborative Research and Development Grant (CRD) – 3D Mobility Mapping Artificial Intelligence (3DM-MAI) and Teledyne Geospatial Inc. We’d like to thank Leihan Chen (Research Scientist), Andrew Sit (Product Manager), Burns Forster (Innovation Manager) and Chris Verheggen (SVP R\& D).

\ifCLASSOPTIONcaptionsoff
  \newpage
\fi



\bibliographystyle{IEEEtran}
\bibliography{bibtex/bib/ref.bib}
%



%




\end{document}